\documentclass{article} 
\usepackage{iclr2019_conference,times}

\usepackage[utf8]{inputenc} 
\usepackage[T1]{fontenc}    
\usepackage{hyperref}       
\usepackage{url}            
\usepackage{amsfonts}       
\usepackage{nicefrac}       
\usepackage{microtype}      
\usepackage{appendix}
\usepackage{graphicx}
\usepackage{subcaption}

\title{Inferring Javascript types using Graph Neural Networks}

\author{%
  Jessica V. Schrouff\thanks{https://prodo.ai/}\\
  Prodo Tech Ltd\\
  London, United Kingdom \\
  \texttt{jessica@prodo.ai} \\
  \And
  Kai Wohlfahrt\\
  Prodo Tech Ltd\\
  London, United Kingdom \\
  \texttt{kai@prodo.ai} \\
  \And
  Bruno Marnette\\
  Prodo Tech Ltd\\
  London, United Kingdom \\
  \texttt{bruno@prodo.ai} \\
  \And
  Liam Atkinson\\
  Prodo Tech Ltd\\
  London, United Kingdom \\
  \texttt{liam@prodo.ai} \\
}

\iclrfinalcopy 
\begin{document}

\maketitle

\begin{abstract}
The recent use of `Big Code' with state-of-the-art deep learning methods offers promising avenues to ease program source code writing and correction. As a first step towards automatic code repair, we implemented a graph neural network model that predicts token types for Javascript programs. The predictions achieve an accuracy above $90\%$, which improves on previous similar work.
\end{abstract}

\section{Introduction}
\label{Intro}
Automatic bug detection or program correction are active fields of research, with recent developments at the intersection of programming languages and machine learning. These approaches have proven useful in the context of dynamic programming languages for which safeguards such as static types are lacking \citep{xu2016}. For instance, in Python or Javascript, inferring the types of code tokens can be challenging and lead to undetected errors. Inferring token types and potentially notifying of type mismatches in a program would hence help preventing undesirable behaviour.

Recent advances in the field of \textit{Big Code} have allowed to e.g. perform code or comment completion, detect code defects or automatically determine token properties, with encouraging success \citep{allamanis2018}. Modelling large amounts of program code is hence a promising avenue to perform automatic source code fixing or provide recommendations during code writing.

In the present work, we use state-of-the-art modelling strategies of program source code to infer Javascript types. Our approach outperforms the current baseline in the field and can easily be deployed in e.g. an online editor or code review tool.

\subsection{Related work}
In \citep{allamanis2017a}, C\# code is represented as a graph including syntactic and semantic information to predict variable names or misuse. While this work provides an interesting framework using graph neural networks and convincing results, the type of each token is used as a feature, which can only be obtained for statically typed languages. On the other hand, previous work on inferring types (e.g. \citep{raychev2015}) did not make use of novel deep learning methods and focused on specific subtasks (e.g. predicting types for function parameters only). While their results are promising, deep learning, and especially Graph Neural Network (GNN, \citep{scarselli2009}) models present the advantages of being typically high accuracy and agnostic models (i.e. the level of pre or post-processing is low). In this work, we represent each Javascript source code as a graph and use state-of-the-art graph neural networks to predict from eight Javascript types for each token.

\section{Materials and methods}
\subsection{Data and code representation}
\label{Data}
Our dataset consists of Javascript source files gathered from Github, excluding files that significantly overlapped\footnote{Based on a similarity analysis as performed using \url{https://www.harukizaemon.com/simian/}}. The dataset is then split once into train and test partitions according to repositories. The training set includes $10,268$ files from $660$ repositories and $578$ organizations, while the testing set includes $3,539$ files from $136$ repositories and $127$ organizations.

We represent each program source code as a graph including syntactic and semantic edges. More specifically, we identify the nodes of the graph\footnote{Please note that we exclude graphs with over $20,000$ nodes (i.e. 4 files) for computational reasons.} from the tokens in the abstract syntax tree (AST, min. nodes = $3$, max. = $19,516$), each node having a feature vector including:

\begin{itemize}
  \item \textit{ast-node-type}: derived from the program's grammar, e.g. `DeclareClass'. We considered 144 possible AST node types (see Appendix \ref{app_datarep}).
  \item \textit{property}: some types have an associated property, e.g. nodes of type `AssignementExpression' can have the property `operator:='. 107 properties were considered.
  \item \textit{values}: each leaf node of the AST has either a name, value or pattern associated. These are stored as strings (capped at 16 characters) for each node.
\end{itemize}

We then construct different types of edges between the obtained nodes based on syntactic (e.g. `child' relationships from the AST) and semantic (e.g. `defined by') relationships between tokens. While our representation differentiates between 96 relationship types (Appendix \ref{app_datarep}), this initial work groups node edges in two categories (AST and reference/traverse).

For each node covered by the test suite of a repository, a label is extracted during run time analysis. These labels correspond to the following JavaScript types: object, string, function, number, undefined, array, boolean and null. Nodes that cannot be associated to a type are assigned an `unknown' type. These are masked out during training to avoid biasing our model towards predicting `unknown'. In addition, nodes with an explicit type-label correspondence (e.g. `DeclareFunction' type and `function' label) are included during the training phase, but excluded from the evaluation phase for a more objective assessment of model performance.

\subsection{Neural Network modelling}

\textbf{Node embeddings:} The node vectors for each feature (i.e. type, property and string) are forwarded to specific embedding layers. For the node types, an embedding of size 144 to size \textit{hidden-size} is used, as nodes can have multiple types. The node properties are encoded as a one-hot vector of size 107 then passed through a linear layer of size \textit{hidden-size}. The strings (of fixed size 16, padded with white space if needed) are first embedded in a lookup table from size 104 symbols to \textit{hidden-size} and then passed through a Gated Recurrent Unit (GRU) of size \textit{hidden-size}, \textit{hidden-size} being a hyper-parameter fixed across all model layers. The different embeddings are then concatenated, before being passed through a number of feed forward neural network blocks consisting of batch-norm normalization, linear layer with dropout and ReLU non-linearity.

\textbf{Graph NNs:} We consider two graph neural network models: the Graph Convolutional Network (GCN, \citep{kipf2016, berg2017}) and the Gated Graph Neural Network (GGNN, \citep{li2015, gilmer2017}).

In the GCN model, a message passing layer collects the messages from a node's neighbours for each edge category (separately), before passing them into a linear layer of size \textit{hidden-size} with dropout. The messages from different edge categories are then summed and passed through a ReLU non-linearity. The number of message passing layers is treated as a hyper-parameter, with weights and dropout untied.

The GGNN model uses a GRU cell to update the node states in each message passing layer. Weights in the message passing phase are tied across layers and messages from different edge categories are passed through a specific linear layer then added together before GRU update, as in \citep{gilmer2017}. Dropout is implemented in the message passing, in which random parts of the messages and of the node vector are dropped before GRU update. The dropout masks are identical across layers, following \citep{gal2016}. In addition, a `master node' is considered for each graph, on which all nodes can write and from which all nodes can read. This master node allows long-distance information to be shared across nodes \citep{gilmer2017}. To this setting, we add a dropout parameter on the nodes that can read or write from the master node, hence creating random `skip connections' across the graph. This dropout value, the dimension of the master node and whether or not to include a master node are considered as hyper-parameters.

\textbf{Decoding}: The decoder, identical for all networks, consists of a log softmax layer preceded by a number of feed forward neural network blocks, as previously described.

\subsubsection{Implementation}
The data is split in mini-batches of 50 graphs or 20,000 nodes (whichever happens first during the random sampling) before entering the embedding, message passing and decoding steps. Accuracy is reported as the F1-score, micro-averaged over the batches in the validation set or over the test set (not batched). The ADAM optimizer \citep{kingma2014} updates the model parameters, with an initial learning rate as manually selected from section \ref{app_lr}. The models include a learning rate scheduler, decreasing the learning rate by a factor of $0.1$ when encountering a plateau in validation accuracy. The best configuration for each model is selected based on a hyper-parameter search, as described in section \ref{app_HS}. All data representations and models were implemented in Python 3.6 using Pytorch 1.0.0.

\section{Results}
\label{Results}
\begin{figure}[!ht]
  \centering
  \includegraphics[height=12cm,width=\linewidth]{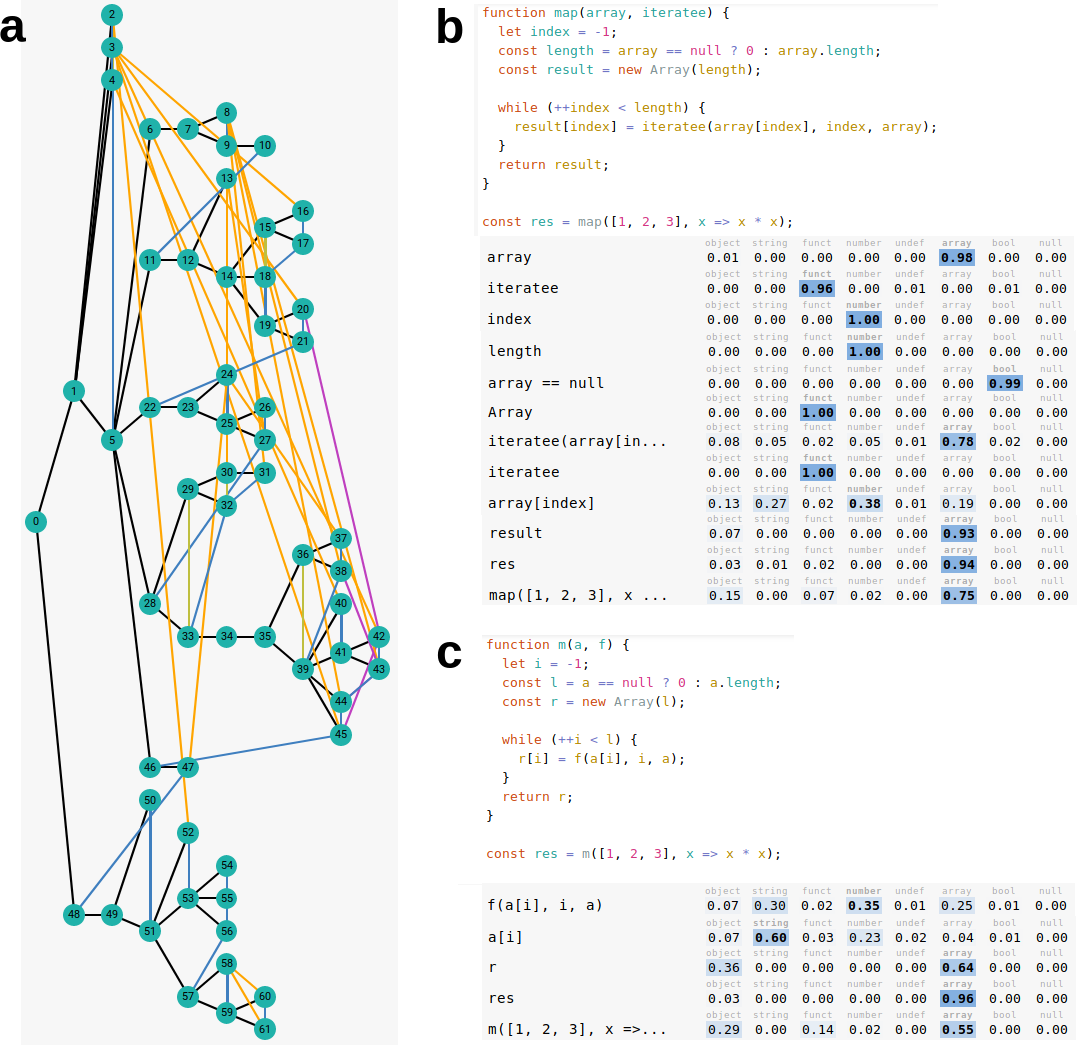}
  \caption{\textbf{a} Graph obtained from the \textit{map} function displayed in \textbf{b}. Black lines represent AST child edges, blue represent `next-node' edges, orange represent `defined-by' edges, magenta represent `next-use' edges and dark yellow represent `next-sibling' edges. \textbf{b} Code with output type predictions (truncated). The model correctly identifies the variables `array', `iteratee' and `result' and can differentiate `Array' from `array'. It however attributes the type `array' to the return of the call to `iteratee', which could be of any type. On the other hand, it is uncertain of its prediction for `array[index]', which could also be of any type. \textbf{c} Same code but removing information from the variable names. In this case, most variables are still correctly identified but `a[i]' is now predicted with some confidence as a string. The model is less confident about its predictions for `r' and `m([1,2,3, x=> x*x])'.}
  \label{fig_results_ill}
\end{figure}

The final results are displayed in Table \ref{results_table} in terms of F1-score and illustrated in Figure \ref{fig_results_ill}. The test set includes a total of $1,043,333$ tokens, among which $751,824$ tokens do not have a ground truth type and $64,631$ have explicit types, hence evaluation is performed on $226,878$ nodes.

\begin{table}[!h]
  \caption{Model performance (in \%).}
  \label{results_table}
  \centering
  \begin{tabular}{lccc}
          & Training & Validation & Test \\
    \\ \hline \\
   	GCN  &  95.04  & 87.69 & 87.25  \\
    GGNN      & 98.01   & 90.52 &  90.79 \\
  \end{tabular}
\end{table}

Both models reach high performance on the considered test set, significantly higher than the 81 \% reported in \citep{raychev2015}. We can further observe that the models perform well on all types apart from `Null' which is less represented in the learning set, and are able to identify missing types on trained node types reasonably well (visual inspection through our web app, demonstrated during the workshop).

\section{Discussion and future work}
\label{Discussion}

This work provides an example of using \textit{Big Code} to infer token types in Javascript source code. Interestingly, both models performed well on this dataset, with only marginal improvement observed for GGNN. This high accuracy could be related to the simplicity of the task, i.e. predicting pre-defined and general types. This aspect limits the potential comparison with the work of \citet{raychev2015}, as they predicted more types, but for a limited number of tokens (i.e. function parameters only). We could extend our label set by adding more general types (e.g. `RegExp' or `Date') or by fine-tuning a model to take into account user specific types, potentially within a project or an organization. 

Future work will investigate the effect of increasing the number of edge type categories on model performance, both in terms of accuracy and computational expenses. In addition, augmenting the semantic information might improve performance, for example `return' relationships were shown to increase accuracy in \citep{allamanis2017a}.

Finally, the inferred types will be used to automatically detect \textit{type mismatches}, e.g. notifying the programmer of potential issues in variable usage. In this context, obtaining uncertainty measures on the type predictions will be beneficial. This could be performed by modifying the dropout implementation \citep{gal2015}. Our next implementation will also take semantic rule into accounts, a common strategy to improve performance and include domain knowledge \citep{raychev2015}.

\section{Data availability}
The data set of this application will be released on our website \url{https://prodo.ai}, as soon as possible. 

\section*{Acknowledgments}
We thank all the team at Prodo Tech for their help on this project, especially Sergio Giro, Mani Sarkar and Jake Runzer.

\bibliography{Pred_Types.bib}

\begin{thebibliography}{15}
\providecommand{\natexlab}[1]{#1}
\providecommand{\url}[1]{\texttt{#1}}
\expandafter\ifx\csname urlstyle\endcsname\relax
  \providecommand{\doi}[1]{doi: #1}\else
  \providecommand{\doi}{doi: \begingroup \urlstyle{rm}\Url}\fi

\bibitem[Allamanis et~al.(2017)Allamanis, Brockschmidt, and
  Khademi]{allamanis2017a}
Miltiadis Allamanis, Marc Brockschmidt, and Mahmoud Khademi.
\newblock Learning to {{Represent Programs}} with {{Graphs}}.
\newblock \emph{arXiv:1711.00740 [cs]}, November 2017.

\bibitem[Allamanis et~al.(2018)Allamanis, Barr, Devanbu, and
  Sutton]{allamanis2018}
Miltiadis Allamanis, Earl~T. Barr, Premkumar Devanbu, and Charles Sutton.
\newblock A {{Survey}} of {{Machine Learning}} for {{Big Code}} and
  {{Naturalness}}.
\newblock \emph{ACM Computing Surveys}, 51\penalty0 (4):\penalty0 81, 2018.

\bibitem[Gal \& Ghahramani(2015)Gal and Ghahramani]{gal2015}
Yarin Gal and Zoubin Ghahramani.
\newblock Dropout as a {{Bayesian Approximation}}: {{Representing Model
  Uncertainty}} in {{Deep Learning}}.
\newblock \emph{arXiv:1506.02142 [cs, stat]}, June 2015.

\bibitem[Gal \& Ghahramani(2016)Gal and Ghahramani]{gal2016}
Yarin Gal and Zoubin Ghahramani.
\newblock A {{Theoretically Grounded Application}} of {{Dropout}} in
  {{Recurrent Neural Networks}}.
\newblock \emph{Proceedings of the 30th Conference on Neural Information
  Processing Systems}, pp.\ ~9, 2016.

\bibitem[Gilmer et~al.(2017)Gilmer, Schoenholz, Riley, Vinyals, and
  Dahl]{gilmer2017}
Justin Gilmer, Samuel~S. Schoenholz, Patrick~F. Riley, Oriol Vinyals, and
  George~E. Dahl.
\newblock Neural {{Message Passing}} for {{Quantum Chemistry}}.
\newblock \emph{arXiv:1704.01212 [cs]}, April 2017.

\bibitem[Kingma \& Ba(2014)Kingma and Ba]{kingma2014}
Diederik~P. Kingma and Jimmy Ba.
\newblock Adam: {{A Method}} for {{Stochastic Optimization}}.
\newblock In \emph{Proceedings of the {{ICLR}} 2015}, December 2014.

\bibitem[Kipf \& Welling(2016)Kipf and Welling]{kipf2016}
Thomas~N. Kipf and Max Welling.
\newblock Semi-{{Supervised Classification}} with {{Graph Convolutional
  Networks}}.
\newblock \emph{arXiv:1609.02907 [cs, stat]}, September 2016.

\bibitem[Li et~al.(2018)Li, Jamieson, Rostamizadeh, Gonina, Hardt, Recht, and
  Talwalkar]{li2018}
Lisha Li, Kevin Jamieson, Afshin Rostamizadeh, Katya Gonina, Moritz Hardt,
  Benjamin Recht, and Ameet Talwalkar.
\newblock Massively {{Parallel Hyperparameter Tuning}}.
\newblock February 2018.

\bibitem[Li et~al.(2015)Li, Tarlow, Brockschmidt, and Zemel]{li2015}
Yujia Li, Daniel Tarlow, Marc Brockschmidt, and Richard Zemel.
\newblock Gated {{Graph Sequence Neural Networks}}.
\newblock \emph{arXiv:1511.05493 [cs, stat]}, November 2015.

\bibitem[Liaw et~al.(2018)Liaw, Liang, Nishihara, Moritz, Gonzalez, and
  Stoica]{liaw2018}
Richard Liaw, Eric Liang, Robert Nishihara, Philipp Moritz, Joseph~E. Gonzalez,
  and Ion Stoica.
\newblock Tune: {{A Research Platform}} for {{Distributed Model Selection}} and
  {{Training}}.
\newblock \emph{arXiv:1807.05118 [cs, stat]}, July 2018.

\bibitem[Raychev et~al.(2015)Raychev, Vechev, and Krause]{raychev2015}
Veselin Raychev, Martin Vechev, and Andreas Krause.
\newblock Predicting {{Program Properties}} from "{{Big Code}}".
\newblock In \emph{Proceedings of the 42nd {{Annual ACM SIGPLAN}}-{{SIGACT
  Symposium}} on {{Principles}} of {{Programming Languages}} - {{POPL}} '15},
  pp.\  111--124, Mumbai, India, 2015. {ACM Press}.
\newblock ISBN 978-1-4503-3300-9.
\newblock \doi{10.1145/2676726.2677009}.

\bibitem[Scarselli et~al.(2009)Scarselli, Gori, Tsoi, Hagenbuchner, and
  Monfardini]{scarselli2009}
F.~Scarselli, M.~Gori, A.~C. Tsoi, M.~Hagenbuchner, and G.~Monfardini.
\newblock The {{Graph Neural Network Model}}.
\newblock \emph{IEEE Transactions on Neural Networks}, 20\penalty0
  (1):\penalty0 61--80, January 2009.
\newblock ISSN 1045-9227.
\newblock \doi{10.1109/TNN.2008.2005605}.

\bibitem[Smith(2018)]{smith2018}
Leslie~N. Smith.
\newblock A disciplined approach to neural network hyper-parameters: {{Part}} 1
  -- learning rate, batch size, momentum, and weight decay.
\newblock \emph{arXiv:1803.09820 [cs, stat]}, March 2018.

\bibitem[van~den Berg et~al.(2017)van~den Berg, Kipf, and Welling]{berg2017}
Rianne van~den Berg, Thomas~N. Kipf, and Max Welling.
\newblock Graph {{Convolutional Matrix Completion}}.
\newblock \emph{arXiv:1706.02263 [cs, stat]}, June 2017.

\bibitem[Xu et~al.(2016)Xu, Zhang, Chen, Pei, and Xu]{xu2016}
Zhaogui Xu, Xiangyu Zhang, Lin Chen, Kexin Pei, and Baowen Xu.
\newblock Python {{Probabilistic Type Inference}} with {{Natural Language
  Support}}.
\newblock In \emph{Proceedings of the 2016 24th {{ACM SIGSOFT International
  Symposium}} on {{Foundations}} of {{Software Engineering}}}, {{FSE}} 2016,
  pp.\  607--618, New York, NY, USA, 2016. {ACM}.
\newblock ISBN 978-1-4503-4218-6.
\newblock \doi{10.1145/2950290.2950343}.

\end{thebibliography}

\appendix
\section{Data representation}
\label{app_datarep}

\begin{verbatim}
{
  "nodeTypes": [
    "AnyTypeAnnotation",
    "ArrayExpression",
    "ArrayPattern",
    "ArrayTypeAnnotation",
    "ArrowFunctionExpression",
    "AssignmentExpression",
    "AssignmentPattern",
    "AwaitExpression",
    "BinaryExpression",
    "BlockStatement",
    "BooleanLiteralTypeAnnotation",
    "BooleanTypeAnnotation",
    "BreakStatement",
    "CallExpression",
    "CatchClause",
    "ClassBody",
    "ClassDeclaration",
    "ClassExpression",
    "ClassImplements",
    "ClassProperty",
    "ConditionalExpression",
    "ContinueStatement",
    "DebuggerStatement",
    "DeclareClass",
    "DeclareExportDeclaration",
    "DeclareFunction",
    "DeclareInterface",
    "DeclareModule",
    "DeclareModuleExports",
    "DeclareOpaqueType",
    "DeclareTypeAlias",
    "DeclareVariable",
    "Decorator",
    "DoExpression",
    "DoWhileStatement",
    "EmptyStatement",
    "EmptyTypeAnnotation",
    "ExistentialTypeParam",
    "ExportAllDeclaration",
    "ExportDefaultDeclaration",
    "ExportDefaultSpecifier",
    "ExportNamedDeclaration",
    "ExportNamespaceSpecifier",
    "ExportSpecifier",
    "ExpressionStatement",
    "ForAwaitStatement",
    "ForInStatement",
    "ForOfStatement",
    "ForStatement",
    "FunctionDeclaration",
    "FunctionExpression",
    "FunctionTypeAnnotation",
    "FunctionTypeParam",
    "GenericTypeAnnotation",
    "Identifier",
    "IfStatement",
    "Import",
    "ImportDeclaration",
    "ImportDefaultSpecifier",
    "ImportNamespaceSpecifier",
    "ImportSpecifier",
    "InterfaceDeclaration",
    "InterfaceExtends",
    "IntersectionTypeAnnotation",
    "JSXAttribute",
    "JSXClosingElement",
    "JSXElement",
    "JSXEmptyExpression",
    "JSXExpressionContainer",
    "JSXIdentifier",
    "JSXMemberExpression",
    "JSXNamespacedName",
    "JSXOpeningElement",
    "JSXSpreadAttribute",
    "JSXSpreadChild",
    "JSXText",
    "LabeledStatement",
    "Literal",
    "LogicalExpression",
    "MemberExpression",
    "MetaProperty",
    "MethodDefinition",
    "MixedTypeAnnotation",
    "NewExpression",
    "NullableTypeAnnotation",
    "NullLiteralTypeAnnotation",
    "NumberTypeAnnotation",
    "NumericLiteralTypeAnnotation",
    "ObjectExpression",
    "ObjectPattern",
    "ObjectTypeAnnotation",
    "ObjectTypeCallProperty",
    "ObjectTypeIndexer",
    "ObjectTypeProperty",
    "ObjectTypeSpreadProperty",
    "OpaqueType",
    "Program",
    "Property",
    "QualifiedTypeIdentifier",
    "RestElement",
    "RestProperty",
    "ReturnStatement",
    "SequenceExpression",
    "SpreadElement",
    "SpreadProperty",
    "StringLiteralTypeAnnotation",
    "StringTypeAnnotation",
    "Super",
    "SwitchCase",
    "SwitchStatement",
    "TaggedTemplateExpression",
    "TemplateElement",
    "TemplateLiteral",
    "ThisExpression",
    "ThisTypeAnnotation",
    "ThrowStatement",
    "TryStatement",
    "TupleTypeAnnotation",
    "TypeAlias",
    "TypeAnnotation",
    "TypeCastExpression",
    "TypeofTypeAnnotation",
    "TypeParameter",
    "TypeParameterDeclaration",
    "TypeParameterInstantiation",
    "UnaryExpression",
    "UnionTypeAnnotation",
    "UpdateExpression",
    "VariableDeclaration",
    "VariableDeclarator",
    "VoidTypeAnnotation",
    "WhileStatement",
    "YieldExpression",
    "BindExpression",
    "ObjectProperty",
    "StringLiteral",
    "NumericLiteral",
    "NullLiteral",
    "BooleanLiteral",
    "Directive",
    "DirectiveLiteral",
    "ObjectMethod",
    "RegExpLiteral",
    "ClassMethod"
  ],
    "propTypes": [
    "{async:true}",
    "{computed:true}",
    "{delegate:true}",
    "{exact:true}",
    "{exportKind:type}",
    "{exportKind:value}",
    "{expression:true}",
    "{generator:true}",
    "{importKind:type}",
    "{importKind:typeof}",
    "{importKind:value}",
    "{kind:const}",
    "{kind:constructor}",
    "{kind:get}",
    "{kind:init}",
    "{kind:let}",
    "{kind:method}",
    "{kind:set}",
    "{kind:var}",
    "{method:true}",
    "{operator:--}",
    "{operator:-}",
    "{operator:-=}",
    "{operator:!}",
    "{operator:!=}",
    "{operator:!==}",
    "{operator:*}",
    "{operator:**}",
    "{operator:**=}",
    "{operator:*=}",
    "{operator:/}",
    "{operator:/=}",
    "{operator:&}",
    "{operator:&&}",
    "{operator:&=}",
    "{operator:%}",
    "{operator:%*}",
    "{operator:%*=}",
    "{operator:%=}",
    "{operator:^}",
    "{operator:^=}",
    "{operator:+}",
    "{operator:++}",
    "{operator:+=}",
    "{operator:<}",
    "{operator:<<}",
    "{operator:<<=}",
    "{operator:<=}",
    "{operator:=}",
    "{operator:==}",
    "{operator:===}",
    "{operator:>}",
    "{operator:>=}",
    "{operator:>>}",
    "{operator:>>=}",
    "{operator:>>>}",
    "{operator:>>>=}",
    "{operator:|}",
    "{operator:|=}",
    "{operator:||}",
    "{operator:~}",
    "{operator:in}",
    "{operator:instanceof}",
    "{operator:typeof}",
    "{operator:void}",
    "{optional:true}",
    "{prefix:true}",
    "{selfClosing:true}",
    "{shorthand:true}",
    "{sourceType:module}",
    "{static:true}",
    "{tail:true}",
    "{value:true}",
    "{variance:minus}",
    "{variance:plus}",
    "{operator:delete}",
    "{flags:}",
    "{flags:g}",
    "{flags:i}",
    "{flags:gi}",
    "{flags:gm}",
    "{flags:m}",
    "{flags:ig}",
    "{flags:mg}",
    "{flags:mi}",
    "{flags:im}",
    "{flags:u}",
    "{flags:y}",
    "{flags:mig}",
    "{flags:s}",
    "{flags:sm}",
    "{flags:iu}",
    "{flags:ug}",
    "{flags:yg}",
    "{flags:my}",
    "{flags:su}",
    "{flags:sum}",
    "{flags:gim}",
    "{flags:gmi}",
    "{flags:iy}",
    "{flags:iyg}",
    "{flags:um}",
    "{flags:iug}",
    "{flags:is}",
    "{flags:sg}",
    "{flags:sy}",
    "{flags:ms}"
  ],
  "edgeTypes": [
    "ast.child.alternate",
    "ast.child.argument",
    "ast.child.arguments",
    "ast.child.attributes",
    "ast.child.block",
    "ast.child.body",
    "ast.child.bound",
    "ast.child.callee",
    "ast.child.callProperties",
    "ast.child.cases",
    "ast.child.children",
    "ast.child.closingElement",
    "ast.child.consequent",
    "ast.child.declaration",
    "ast.child.declarations",
    "ast.child.decorators",
    "ast.child.discriminant",
    "ast.child.elements",
    "ast.child.elementType",
    "ast.child.exported",
    "ast.child.expression",
    "ast.child.expressions",
    "ast.child.extends",
    "ast.child.finalizer",
    "ast.child.handler",
    "ast.child.id",
    "ast.child.implements",
    "ast.child.impltype",
    "ast.child.imported",
    "ast.child.indexers",
    "ast.child.init",
    "ast.child.key",
    "ast.child.label",
    "ast.child.left",
    "ast.child.local",
    "ast.child.meta",
    "ast.child.name",
    "ast.child.namespace",
    "ast.child.object",
    "ast.child.openingElement",
    "ast.child.param",
    "ast.child.params",
    "ast.child.properties",
    "ast.child.property",
    "ast.child.qualification",
    "ast.child.quasi",
    "ast.child.quasis",
    "ast.child.rest",
    "ast.child.returnType",
    "ast.child.right",
    "ast.child.source",
    "ast.child.specifiers",
    "ast.child.superClass",
    "ast.child.supertype",
    "ast.child.superTypeParameters",
    "ast.child.tag",
    "ast.child.test",
    "ast.child.typeAnnotation",
    "ast.child.typeParameters",
    "ast.child.types",
    "ast.child.update",
    "ast.child.value",
    "ast.child",
    "ast.next-in-list.arguments",
    "ast.next-in-list.attributes",
    "ast.next-in-list.body",
    "ast.next-in-list.callProperties",
    "ast.next-in-list.cases",
    "ast.next-in-list.children",
    "ast.next-in-list.consequent",
    "ast.next-in-list.declarations",
    "ast.next-in-list.decorators",
    "ast.next-in-list.elements",
    "ast.next-in-list.expressions",
    "ast.next-in-list.extends",
    "ast.next-in-list.implements",
    "ast.next-in-list.indexers",
    "ast.next-in-list.params",
    "ast.next-in-list.properties",
    "ast.next-in-list.quasis",
    "ast.next-in-list.specifiers",
    "ast.next-in-list.types",
    "ast.next-in-list",
    "ast.position-from-last.*",
    "ast.position-from-last.1",
    "ast.position-from-last.2",
    "ast.position-from-last.3",
    "ast.position-from-last.4",
    "ast.position.*",
    "ast.position.0",
    "ast.position.1",
    "ast.position.2",
    "ast.position.3",
    "ast.position.4",
    "reference.defined-by",
    "reference.next-use",
    "traverse.next-node",
    "traverse.next-sibling",
    "ast.child.directives",
    "ast.next-in-list.directives"
  ],
}
\end{verbatim}

\section{Hyper-parameter setting}
We split the training set (based on repositories) into a single 80\% - 20\% partition (fixed random seed) to perform hyper-parameter search.

\subsection{Learning rate}
\label{app_lr}

\begin{figure}[h!]
  \centering
  \begin{subfigure}[b]{0.4\linewidth}
    \includegraphics[width=\linewidth]{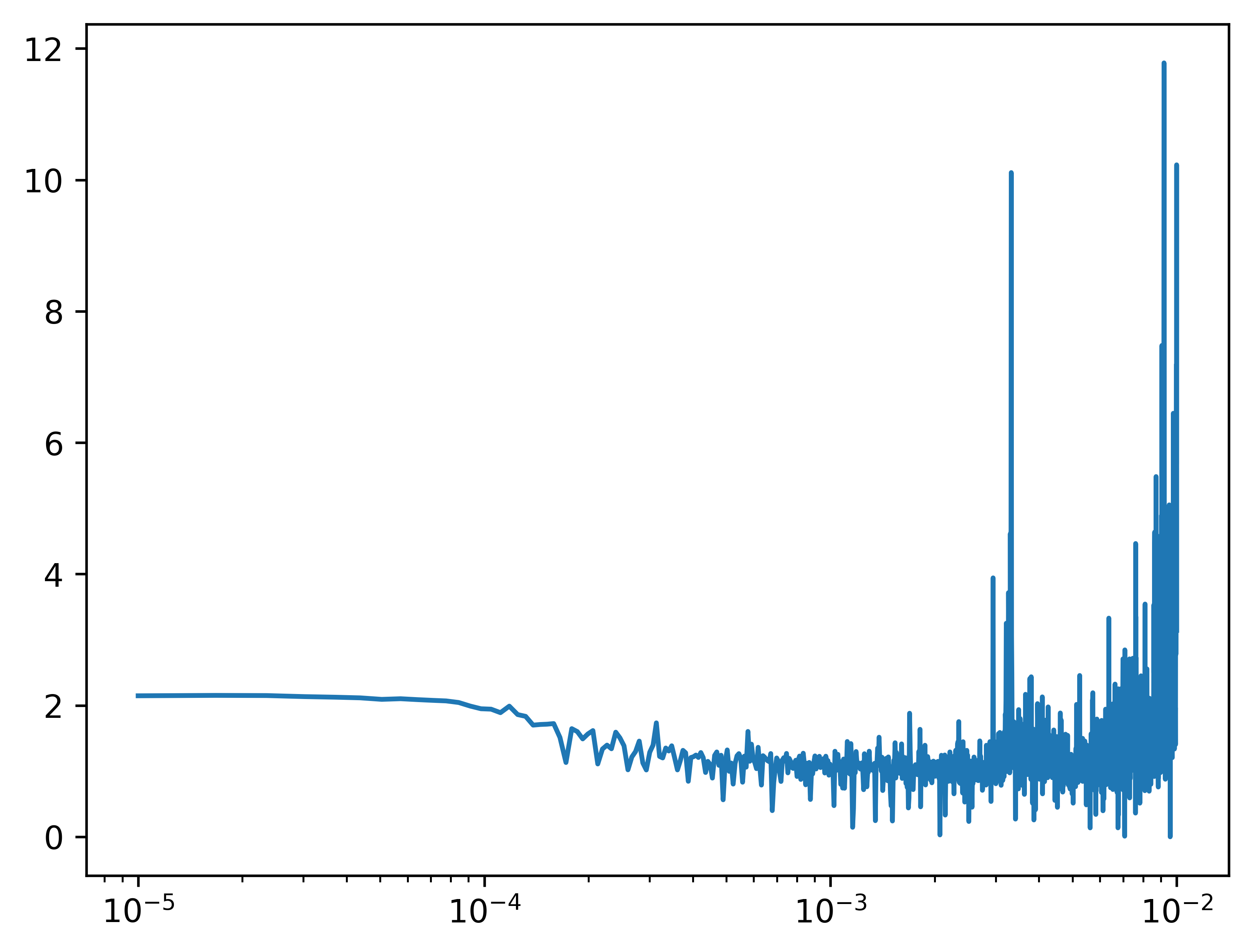}
    \caption{GCN.}
  \end{subfigure}
  \begin{subfigure}[b]{0.4\linewidth}
    \includegraphics[width=\linewidth]{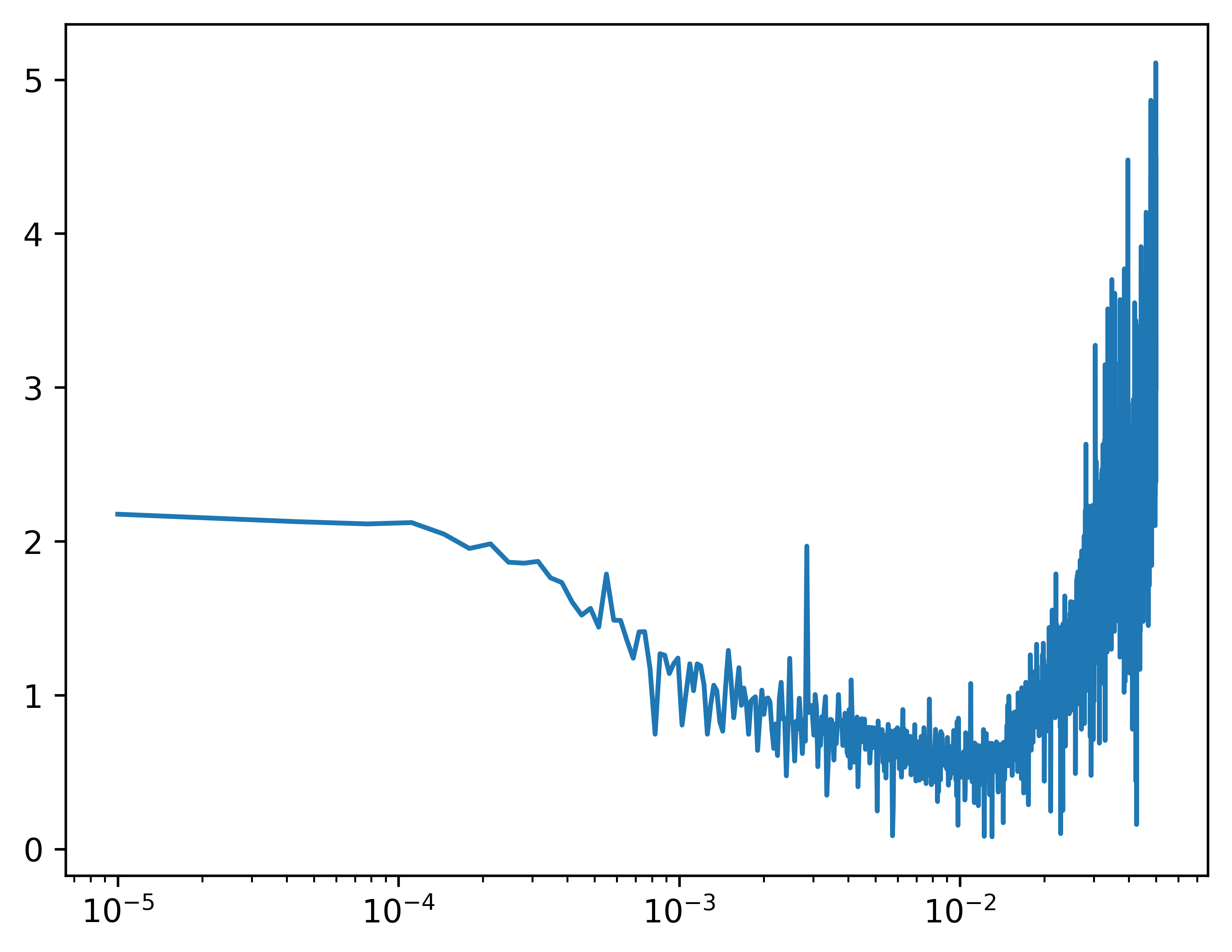}
    \caption{GGNN.}
  \end{subfigure}
  \caption{Learning rate finders for the considered models. The x-axis represents slowly increasing the learning rate after each batch while recording the training loss (y-axis).}
  \label{fig_lr}
\end{figure}

We first use the training set to identify a reasonable starting point for the learning rate of each model, as proposed in \citep{smith2018}. The learning rate finders are presented in Figure \ref{fig_lr}.  These plots display the loss on the training set while slowly increasing the learning rate after each forward pass on a minibatch, across a total of eight epochs. The initial learning rate value is selected at the middle of the `acceptable' range, where the minimum is identified as the moment the model starts learning and the maximum as the moment the curve becomes too rough or training loss is at its minimum.

The selected values are $0.0004$ for GCN and $0.001$ for GGNN.

\subsection{Model hyper-parameters}
\label{app_HS}

This learning rate is then used to search over the space of hyper-parameters identified for each model. More specifically, we use Ray \url{https://ray.readthedocs.io/en/latest/index.html} and Tune \citep{liaw2018} to perform an Asychronous HyperBand search \citep{li2018} with a maximum of 50 epochs. 200 configurations are tested, uniformly sampled from:

\begin{itemize}
\item node representation size (\textit{hidden-size}): 32, 64 or 128
\item number of encoding feed forward blocks: 0 to 3
\item dropout rate (constant across the model): 0, 0.1, 0.2, 0.3, 0.4, 0.5
\item number of message passing steps: 1 to 10
\item number of decoding feed forward blocks: 0 to 3
\end{itemize}

In addition, for GGNN we randomly include a master node of size 20 to 200 by steps of 20 with a dropout rate of 0 to 0.5 by steps of 0.1.

For each model, the validation accuracy for each configuration is extracted and plotted against each value of a hyper-parameter. The plot overlays a box plot and a scatter plot for more insight. It is expected that configurations leading to overall lower performance would be stopped earlier, leading to a decreased average for a specific value of the hyper-parameter. Please note that this plot does not investigate potential interactions between hyper-parameters. In addition, we manually explore the 10 configurations leading to highest performance.

\begin{figure}[h!]
  \centering
  \includegraphics[width=\linewidth]{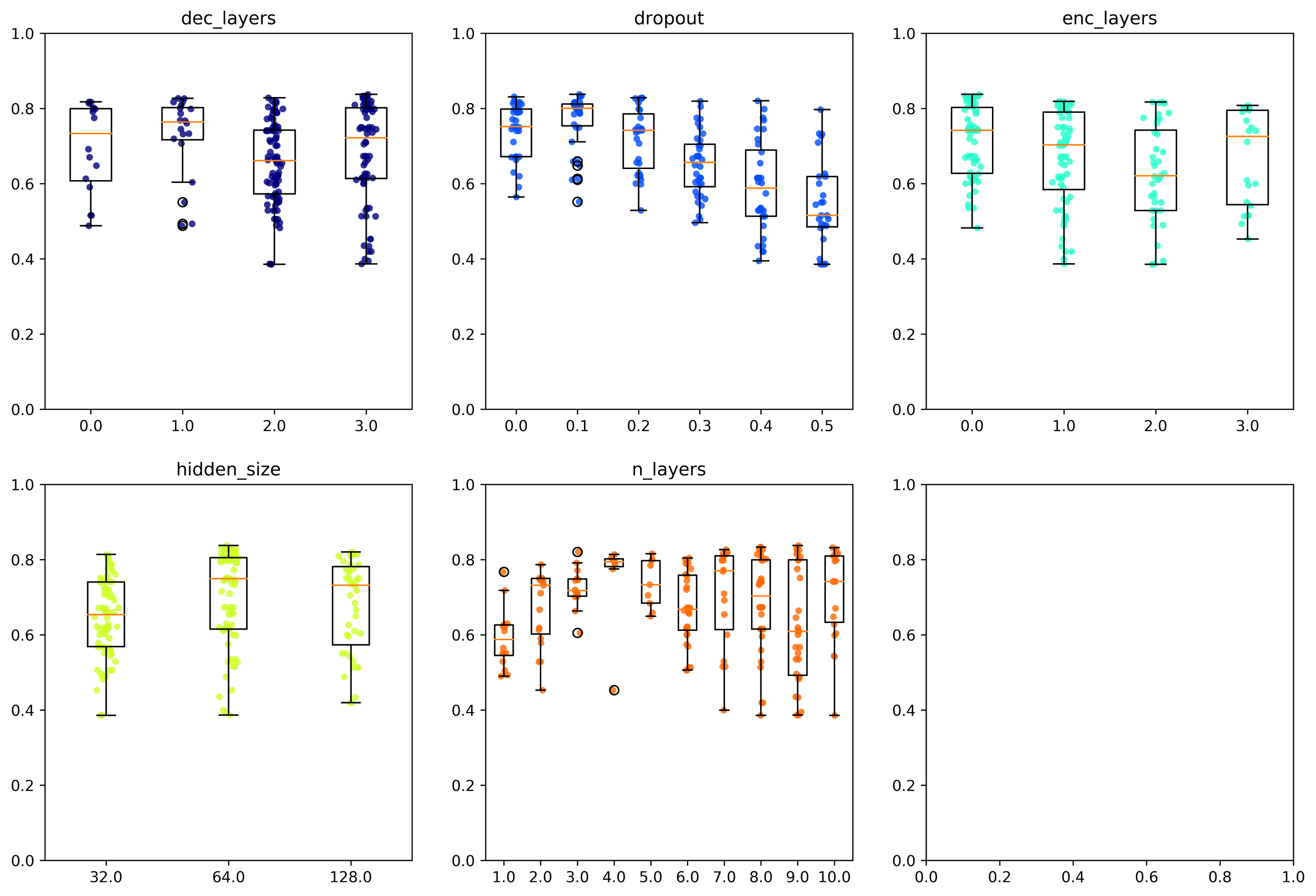}
  \caption{Hyper-parameter search for the GCN model. The space includes the number of encoding layers ($enc_{layers}$), the size of node representation ($hidden_{size}$), the dropout rate ($dropout$), the number of message passing layers ($n_{layers}$) and the number of decoding layers ($dec_{layers}$). This figure is based on 200 successful trials.}
  \label{fig_hyper_gcn}
\end{figure}

These 10 configurations for the GCN model include no encoding layer, a node representation size of 64 and a low but preferably non-null dropout rate (0: 1, 0.1: 4, 0.2: 5). The number of message passings is preferred as high, including between 7 and 10 convolutions. The number of decoding layers is also favoured as high (3: 7, 2: 1, 1: 2). Those results are illustrated in Figure \ref{fig_hyper_gcn}. Based on this figure and the 10 best configurations, a sensible configuration for the GCN model would hence be: no encoding layer, hidden size of 64, dropout of 0.1, 7 layers of message passing and 1 decoding layer.

\begin{figure}[h!]
  \centering
  \includegraphics[width=\linewidth]{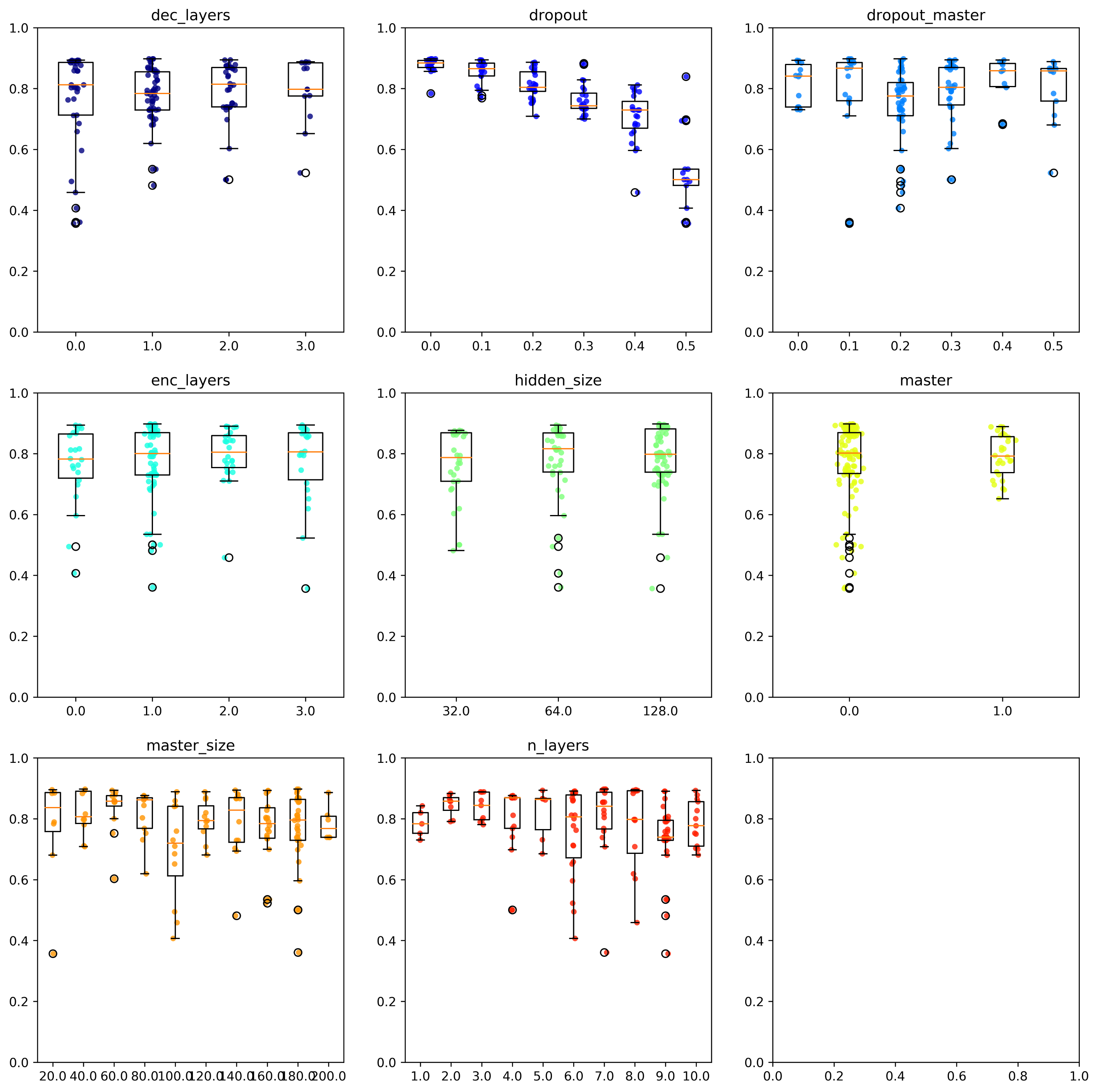}
  \caption{Hyper-parameter search for the GGNN model. The space includes the number of encoding layers ($enc_{layers}$), the size of node representation ($hidden_{size}$), the dropout rate ($dropout$), the number of message passing layers ($n_{layers}$), whether to include a master node ($master$, with $master_{size}$ and $dropout_{master}$) and the number of decoding layers ($dec_{layers}$). This figure is based on 138 successful trials (out of memory errors encountered).}
  \label{fig_hyper_ggnn}
\end{figure}

For the GGNN model, the 10 best configurations include a low number of encoding layers after the embedding (0 layers: 1, 1: 6, 2: 2, 3: 1) and prefer a higher node representation size (128: 9 out of 10, 64: 1) with low dropout (0: 7 out of 10, 0.1: 3). As displayed in Figure \ref{fig_hyper_ggnn}, the number of message passing time steps is variable, with a reasonable choice being between 4 and 8 (8 out of 10). Surprisingly, the master node does not seem to improve classification accuracy (9 out of 10 configurations without master). This result suggests that adding long range connections does not improve on the graph representation. This might be due to the specific tree structure established from the AST that is relatively brittle (i.e. not all permutations can be performed on the graph and it does not make sense to connect all nodes). Alternatively, the model might become too large, overfit and needs more data. Finally, a low number of decoding layers is preferred (0: 4, 1: 4, 2: 2). The configuration chosen for the GGNN includes 1 encoding layer, a hidden size of 128, no dropout, 5 message passing layers without master node and 1 decoding layer.

\end{document}